\documentclass{article}

\usepackage{arxiv}

\usepackage[utf8]{inputenc} 
\usepackage[T1]{fontenc}    
\usepackage{hyperref}       
\usepackage{url}            
\usepackage{booktabs}       
\usepackage{amsfonts}       
\usepackage{nicefrac}       
\usepackage{microtype}      
\usepackage{lipsum}

\usepackage{graphicx}
\usepackage{amssymb}
\usepackage{lineno}
\usepackage{algorithm}
\usepackage{algpseudocode}
\usepackage{amsmath,amssymb,array}
\usepackage{booktabs}
\usepackage{adjustbox,lipsum}
\usepackage{multirow}
\usepackage{soul}
\usepackage{algorithm}
\usepackage{algorithmicx}
\usepackage{algpseudocode}
\usepackage{lscape}
\usepackage{listings}
\usepackage{verbatim}
\usepackage{hyperref}

\title{Supervised learning with artificial hydrocarbon networks: an open source implementation and its applications}

\author{
  Jos\'{e}~Roberto~Ayala-Solares\\
  Amazon\\
  Seattle, WA 98121, USA\\
   \And
 Hiram~Ponce\\
  Facultad de Ingenier\'{i}a\\
  Universidad Panamericana\\
  Ciudad de M\'{e}xico, 03920, M\'{e}xico \\
  \texttt{hponce@up.edu.mx} \\
}

\begin{document}
\maketitle

\begin{abstract}
Artificial hydrocarbon networks (AHN) is a novel supervised learning method inspired on the structure and the inner chemical mechanisms of organic compounds. As any other cutting-edge algorithm, there are two challenges to be faced: time-consuming for encoding and complications to connect with other technologies. Large and open source platforms have proved to be an alternative solution to the latter challenges. In that sense, this paper aims to introduce the \texttt{ahnr} package for R that implements AHN. It provides several functions to create, train, test and visualize AHN. It also includes conventional functions to easily interact with the trained models. For illustration purposes, it presents several examples about the applications of AHN in engineering, as well as, the way to use it. This package is intended to be very useful for scientists and applied researchers interested in machine learning and data modeling. Package availability is in the Comprehensive R Archive Network.
\end{abstract}


\section{Introduction}
\label{sec:introduction}

Artificial hydrocarbon networks (AHNs) is a supervised learning method inspired on the structure and the inner chemical mechanisms of organic compounds \cite{Ponce2011artificial,Ponce2014artificial}. It aims to package information, from a set of instances and in a hierarchical approach, loosely inspired on the way atoms and molecules interact to produce organic, carbon-based networks. As a result, this method performs a supervised learning task that can solve regression and classification problems \cite{Ponce2011artificial,Ponce2014artificial,Ponce2016flexible}. Moreover, this method can handle uncertain and imprecise information typically found in real-world, such as: sales prediction \cite{Ponce2015artificial}, forecasting \cite{Ponce2011artificial,Ponce2014artificial}, signal processing \cite{Ponce2014adaptive,Ponce2014artificial}, engineering control systems \cite{Ponce2013artificial,Ponce2015novel,Ponce2017doubly}, and others. 

However, two challenges are present when facing the introduction of a new method: time-consuming for encoding and complications to connect with other technologies \cite{Ponce2015development}. Even more, high performance and high portability are required in data computational applications. Thus, choosing the appropiate development programming language is not a trivial task \cite{Dogaru2015using}. In that sense, large and open source platforms have proved to be an alternative solution to implement algorithms. For instance, the R language is a good candidate that provides an open source framework, zero-cost software, wide support community, that compiles in a variety of operating systems. Also, it is widely used by scientists and researchers in machine learning and data science \cite{Lantz2015machine}.

In literature, it is found that AHN has been implemented in distinct specialized software. For example, an informal MATLAB implementation of this method has been reported in \cite{Ponce2014artificial}. Also, AHNs have been implemented in the Artificial Organic Networks Toolkit for LabVIEW \cite{Ponce2015development}. However, a free specialized software for AHN has not been done before.

In that sense, this paper aims to introduce the \texttt{ahnr} package for R that implements AHN. This package is the first free specialized software for AHN.  It is intended to consolidate in a single R package an accessible and easy-to-use tool to implement AHN, with functions that facilitate its creation, training, testing and visualization. It also includes conventional functions to easily interact with the trained models in the R environment. 

To illustrate the implementability and applicability of AHN, this work presents several examples about the applications of AHN in engineering, the usage of the \texttt{ahnr} package, and a real implementation for predicting exchange rates for Brazilian, Russian, Indian and Chinese (BRIC) currencies to U.S. dollars. Readers are referred to the \texttt{ahnr} package which is available in the Comprehensive R Archive Network (CRAN): \url{https://cran.r-project.org/web/packages/ahnr/index.html}.

The rest of the paper is organized as follows. Section \ref{sec:ahn} presents an overview of artificial hydrocarbon networks. In Section \ref{sec:pkg}, it is described the implementation of the \texttt{ahnr} package, and two examples are introduced to explain the usage of it. Section \ref{sec:case} presents an application for predicting the exchange rates for BRIC currencies to U.S. dollars using the \texttt{ahnr} package.  Lastly, Section \ref{sec:limitations} provides some limitations of the current implementation of the package, and finally Section \ref{sec:conclusion} concludes the paper.

\section{Artificial Hydrocarbon Networks}
\label{sec:ahn}

Artificial hydrocarbon networks is a supervised learning method. It aims to model data using the inspiration of carbon networks, simulating the chemical rules involved within organic molecules to represent the structure and behavior of data \cite{Ponce2014artificial,Ponce2016novel}. This method inherits from a general framework of learning algorithms namely artificial organic networks \cite{Ponce2014artificial}. The latter proposes to build artificial organic models, including: (i) a graph structure related to their physical properties, and (ii) a mathematical model behavior related to their chemical properties. The main property of artificial organic networks is packaging information in modules called molecules \cite{Ponce2014artificial}. These packages are then organized and optimized using heuristic mechanisms based on chemical energy, defined in the training algorithm. 

Depending on the learning method, the graph structure shapes different topologies, the mathematical model is defined appropriately to that structure, and the training algorithm works to find suitable parameters under that artificial organic model \cite{Ponce2014artificial,Ponce2017interpretability}. More details can be found in \cite{Ponce2014artificial,Ponce2015development}.

\subsection{Description of the AHN-Algorithm}

\begin{figure}[t!]
\begin{centering}
\includegraphics[width=0.5\textwidth]{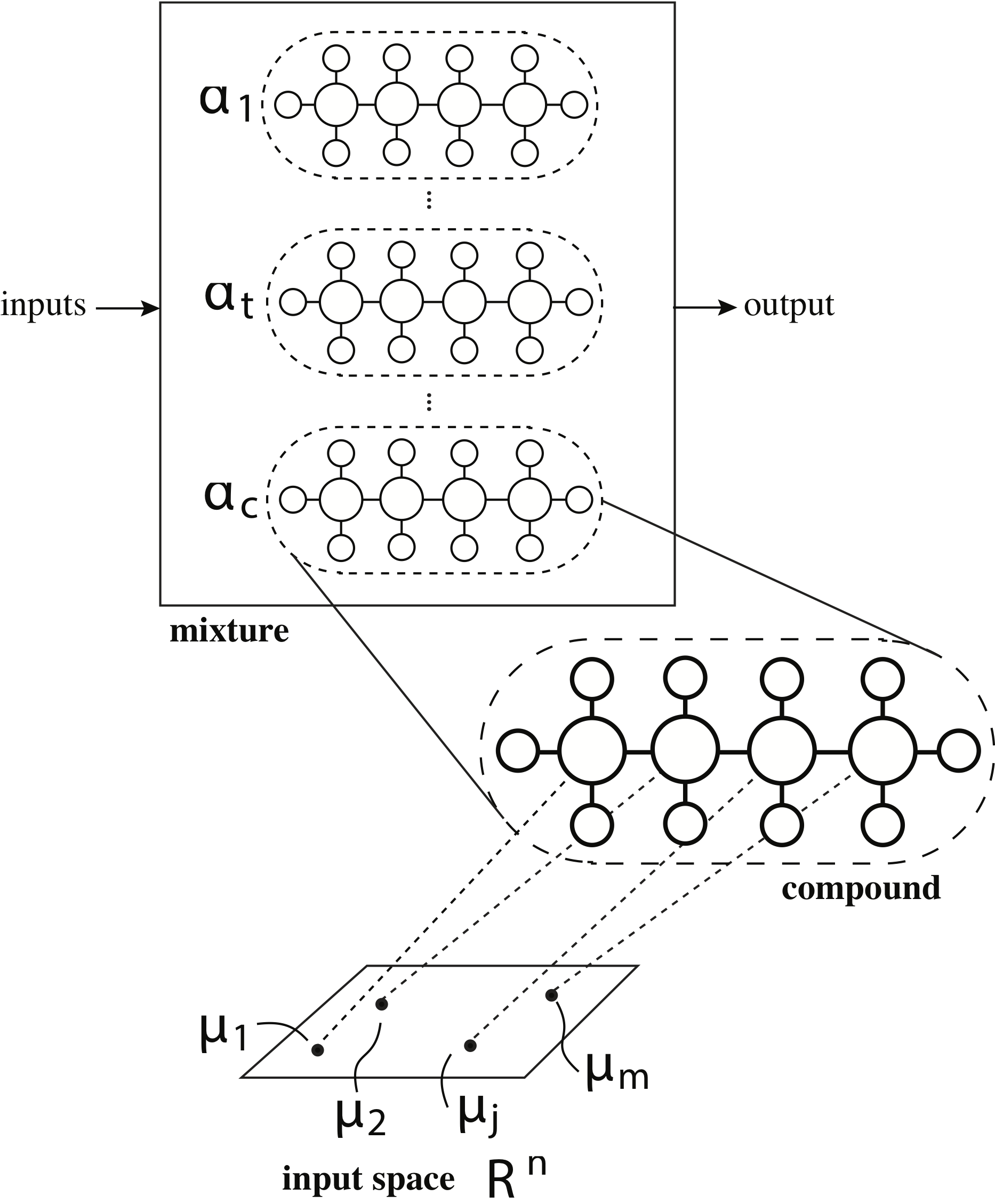}
\par\end{centering}
\caption{Structure of AHN using saturated and
linear chains of molecules.\label{fig:block-diagram-AHN}}
\end{figure}

AHN is a supervised learning method that loosely simulates the chemical interactions of hydrocarbon molecules. For readability, Table \ref{tab:terms} summarizes the description of the chemical-based terms and their meanings used in the AHNs technique introduced below \cite{Ponce2016flexible,Ponce2014artificial}.

\begin{table}
\caption{Description of the chemical terms used in AHN and their computational meanings, adapted from \cite{Ponce2014artificial}.\label{tab:terms}}
\centering
\footnotesize
\begin{tabular}{|l|l|l|} 
\hline \textbf{Chemical terminology} & \textbf{Symbol} & \textbf{Meaning} \\
\hline environment & $x$ & (features) data inputs in $\mathbb{R}^n$ \\
\hline ideal behavior & $y$ & (target) data outputs, solution of mixtures \\
\hline atoms & $H_i$, $\sigma$ & (parameters) basic structural units or properties \\
\hline molecules & $\varphi(x)$ & (functions) basic units of information \\
\hline compounds & $\psi(x)$ & (composite functions) complex units of information made of $m$ molecules \\
\hline mixtures & $S(x)$ & (linear combinations) combination of compounds \\
\hline stoichiometric coefficient & $\alpha_t$ & (weights) definite ratios in mixtures \\
\hline molecular centers & $\mu_j$ & (vector) representation of the centered data in molecules\\
\hline energy & $E_0,E_j$ & (loss function) value of the error between target and estimate values \\
\hline \end{tabular} 
\normalsize
\end{table}

This algorithm is only composed of hydrogen and carbon elements that can be linked together with up to one and four atoms, respectively (see Fig. \ref{fig:block-diagram-AHN}). The basic unit information is called CH-molecule, or simply molecule, and it is formed when one carbon atom is linked with $1 \leq k \leq 4$ hydrogen atoms, denoted as $CH_k$. Generally speaking, one molecule models a chunk of data in its parameters (hydrogen and carbon atoms) and configuration. The molecule has a structural representation (i.e. configuration) and a chemical behavior. Mathematically, the behavior $\varphi$ of a molecule with $k$ hydrogen atoms is expressed as in (\ref{eq:CHmolecule}); where, $\sigma \in \mathbb{R}^n$ is called the carbon value, $H_i \in \mathbb{R}^n$ is the $i$-th hydrogen atom attached to the carbon atom, and $x=(x_1,\dots,x_n)$ is the input vector with $n$ features. 

\begin{equation}
\varphi(x,k)= \sum_{r=1}^n \sigma_r\sum_{i=1}^{k\leq4}H_{ir}x^k
\label{eq:CHmolecule}
\end{equation}

Two or more unsaturated molecules, i.e. those with hydrogen atoms less than 4, can be joined together. In AHN, this new structure is called compound. Different compounds have been defined in literature \cite{Ponce2014artificial,Ponce2015development,Ponce2014adaptive}. The simplest one is the saturated and linear chain of $m$ molecules. It is composed structurally of two $CH_3$ molecules and $(m-2)$ $CH_2$ molecules. The behavior $\psi$ of a saturated-and-linear compound is defined as (\ref{eq:psicompound}); where, $\varphi_j$ is the behavior of the $j$th  associated molecule that represents a subset $\Sigma_j$ of the input $x$ such that $\Sigma_j=\{x|\arg\min_j(x-\mu_j)=j\}$, and $\mu_j \in \mathbb{R}^n$ is the center of the $j$th molecule \cite{Ponce2016flexible,Ponce2017interpretability}. In fact, $\Sigma_{j_1} \cap \Sigma_{j_2} = \emptyset$ if $j_1 \neq j_2$.

\begin{equation}
\psi(x)=\begin{cases}
\varphi_{1}(x,3) & x \in \Sigma_1 \\
\varphi_{2}(x,2) & x \in \Sigma_2 \\
\cdots & \cdots\\
\varphi_{m-1}(x,2) & x \in \Sigma_{m-1} \\
\varphi_{m}(x,3) & x \in \Sigma_m
\end{cases}\label{eq:psicompound}
\end{equation}

Compounds can interact among them in definite ratios $\alpha_t$, so-called stoichiometric coefficients or weights, forming a mixture $S(x)$, expressed as in (\ref{eq:mixture}); where, $c$ represents the number of compounds in the mixture and $\alpha_t$ is the weighted factor of the $t$-th compound \cite{Ponce2014artificial}. 

\begin{equation}
S(x)=\sum_{t=1}^{c}\alpha_{t}\psi_{t}(x)\label{eq:mixture}
\end{equation}

Formally, an AHN is a mixture of compounds (see Fig. \ref{fig:block-diagram-AHN}) each one computed using a chemical-based heuristic rule, expressed in the so-called AHN-algorithm \cite{Ponce2014artificial,Ponce2011artificial,Ponce2015development}. Literature reports extensive usage of AHN with a single saturated-and-linear compound \cite{Ponce2011artificial,Ponce2014artificial,Ponce2014adaptive,Ponce2016flexible,Ponce2016novel,Ponce2017interpretability}. Finally, with the latter restrictions, the AHN-algorithm is computed like Algorithm \ref{alg:AHN-algorithm}.

\subsection{The Simple AHN-Algorithm}
At first, Algorithm \ref{alg:AHN-algorithm} initializes the structure of a saturated and linear compound with $m$ molecules, and the centers of molecules $\{\mu_j\}$ are randomly set. While a stop criterion is not reached, the compound is computed and updated as follows. First, for each molecule, the training dataset $\Sigma$ is partitioned in subsets $\Sigma_j$ such that every input $x$ is near to $\mu_j$. Then, the hydrogen and carbon values of each molecule is computed independently using the least squares estimates (LSE) method, and the error $E_j$ between the output response of the $j$th molecule and the actual targets $y$ of the $j$th subset is calculated. If any of the subsets is empty, then these subsets require a new position, i.e. update their centers. This relocation is made by simply changing the center of the empty subset randomly close to one molecule with large error. Later on, the center of molecules are updated via a gradient descent approach with learning rate $0 < \eta < 1$, previously chosen, as shown in (\ref{eq:center}) with $E_0 = 0$. Lastly, the compound $\psi$ is updated with the behaviors of molecules calculated so far. A detailed description of the AHN-algorithm can be found in \cite{Ponce2014artificial}. 

\begin{equation}
\mu_j \leftarrow \mu_j - \eta (E_{j-1} - E_j) \label{eq:center}
\end{equation}

\begin{algorithm}
\noindent \textbf{\footnotesize Input:}{\footnotesize{} the training
data set $\Sigma=(x,y)$, the number of molecules in the compound
$m \geq 2$ and the learning rate $0< \eta < 1$.}{\footnotesize \par}
\noindent \textbf{\footnotesize Output:}{\footnotesize{} the trained compound $\psi$.}{\footnotesize \par}
\noindent {\footnotesize $\;$}{\footnotesize \par}
\footnotesize{
\begin{algorithmic}[1]
\State {Create the structure of a new compound with $m$ molecules.}
\State {Randomly initialize a set of molecular centers, $\mu_j$.}
\While {not reach a stop criterion}
\For {$j=1$ to $m$}
\State $\Sigma_j \leftarrow$ do a subset of $\Sigma$ using $\mu_j$.
\State $\{H_j,\sigma_j\} \leftarrow$ calculate molecular parameters of $\varphi_j$ using LSE method.
\State $E_j \leftarrow$ compute the error in molecule.
\EndFor
\For {$j=1$ to $m$}
\If {$\Sigma_j = \emptyset$}
\State $\mu_j \leftarrow$ relocate the center of molecule near to other $\mu_k$ such that $E_k$ is very large.
\EndIf
\EndFor
\State Update all centers using $\mu_j = \mu_j - \eta(E_{j-1}-E_j)$.
\EndWhile
\State Update the behavior of compound $\psi$ using all $\varphi_j$ already calculated.\\
\Return $\psi$.
\end{algorithmic}
}
\caption{Training algorithm for a single saturated-and-linear compound of AHN.}
\label{alg:AHN-algorithm}
\end{algorithm}

\subsection{Properties of Artificial Hydrocarbon Networks}
AHN algorithm has several properties very useful when considering regression and classification problems: stability, robustness, packaging data and parameter interpretability. 

Stability implies that the AHN-algorithm minimizes the changes in its output response when the inputs slightly change \cite{Ponce2014artificial}, promoting the usage of the AHNs as a supervised learning method. Robustness considers that the AHN-algorithm can deal with uncertain and noisy data which implies that it behaves as a filtering information system. Literature reports different examples about robustness, as seen in \cite{Ponce2014adaptive,Ponce2013artificial,Ponce2015novel,Ponce2017doubly,Ponce2016novel,Ponce2016flexible}. Furthermore, packaging data enables to compute molecular structures into the algorithm in the sense that similar data with similar capabilities are clustered together. This property intuitively reveals that data is not only packaged by its features, but also by its tendency \cite{Ponce2014artificial}. Lastly, parameter interpretability refers to that AHN can be useful to partially understand underlying information or to extract features. For example, the AHN-algorithm has been used in facial recognition approaches when using its parameters as metadata information \cite{Ponce2014artificial}, or it has been implemented in a cancer diagnosis system to interpret its decision-making tasks \cite{Ponce2017interpretability}.

More information about properties and comparisons with other learning methods can be seen in \cite{Ponce2014artificial,Ponce2016flexible,Ponce2017interpretability}.

\section{The \texttt{ahnr} Package: Implementation, Usage and Examples}
\label{sec:pkg}

The \texttt{ahnr} package has been implemented with several functions that facilitate the creation, training, testing and visualization of AHNs. The implementation of the package and the usage of it are shown following.

\subsection{Implementation of the \texttt{ahnr} Package}

There are four main functions in the \texttt{ahnr} package: \texttt{fit}, \texttt{summary}, \texttt{visualize} and \texttt{predict}.

\begin{itemize}
\item The \texttt{fit} function receives a list of two dataframes, one for the inputs and one for the outputs, respectively. The user should specify the number of molecules to use, the learning rate and the maximum number of iterations.  The values for both the number of molecules and the learning rate can be learned using a hyperparameters search approach like grid search, random search or Bayesian optimization. The output of this function is an object of class \texttt{ahn} that contains: (i) the structure of the AHN trained, (ii) original output variable, (iii) predicted output variable, (iv) learning rate used, (v) minimum error achieved, and (vi) names of the input variables.

\item The \texttt{summary} function receives an \texttt{ahn} object and provides  information about the trained AHN. It  shows  the  centers  of  the  molecules  together  with the corresponding coefficients for the elements in each molecule.

\item The \texttt{visualize} function of the \texttt{ahn} object provides an interactive visualization with the structure of the trained network.

\item The \texttt{predict} function uses a trained AHN and uses it to compute output values given new input data.
\end{itemize}

\subsection{Toy Example for Nonlinear Function Approximation}

Consider a dataset where the input variables are defined as $x_1 = \cos{t}$ and $x_2 = t$, and the output variable is defined as $y = \sin{t}$, where $0 \le t \le 15$. In this case, the purpose is to train a network that is able to map both inputs to the output. First, a list of two dataframes needs to be created in R. Then, we use the \texttt{fit} function with 5 molecules, a learning rate of 0.01 and a maximum number of iterations of 2000. The following code shows an excerpt of this implementation:

\begin{verbatim}
> set.seed(123)
> t <- seq(0, 15, 0.01)
> X <- data.frame(x1 = cos(t), x2 = t)
> Y <- data.frame(y = sin(t))

> Sigma <- list(X = X, Y = Y)

> ahn <- fit(Sigma, 5, 0.01, 2000)
\end{verbatim}

After the AHN is trained, it is possible to inspect the trained model using the \texttt{summary} function:

\begin{verbatim}
> summary(ahn)

Artificial Hydrocarbon Network trained:

Number of molecules:
 5 

Learning factor:
 0.01 

Overall error:
 0.038 

Centers of the molecules:
                  x1         x2
molecule1 -0.4602019 11.7892192
molecule2  0.3725416 14.4995160
molecule3  1.0225322  0.8249451
molecule4  0.6767031 14.0067772
molecule5 -0.7064909  6.0398596

Molecules:
Molecule 1:           Molecule 2:
        x1     x2             x1     x2
C1   0.932  0.932     C2   0.102  0.102
H11  0.148  3.208     H21 -0.722  0.838
H12  0.176 -0.642     H22 -0.399 -0.055
H13 -0.035  0.030

Molecule 3:           Molecule 4:
        x1     x2             x1     x2
C3   0.086  0.086     C4  -0.199 -0.199
H31 -0.004  1.058     H41  1.240 -1.078
H32 -0.174 -0.338     H42 -0.443  0.083

Molecule 5:
        x1     x2
C5   6.395  6.395
H51 -0.747 -9.148
H52  0.320  1.838
H53 -0.152 -0.110
\end{verbatim}

\begin{figure}
\begin{centering}
\includegraphics[width=0.7\textwidth]{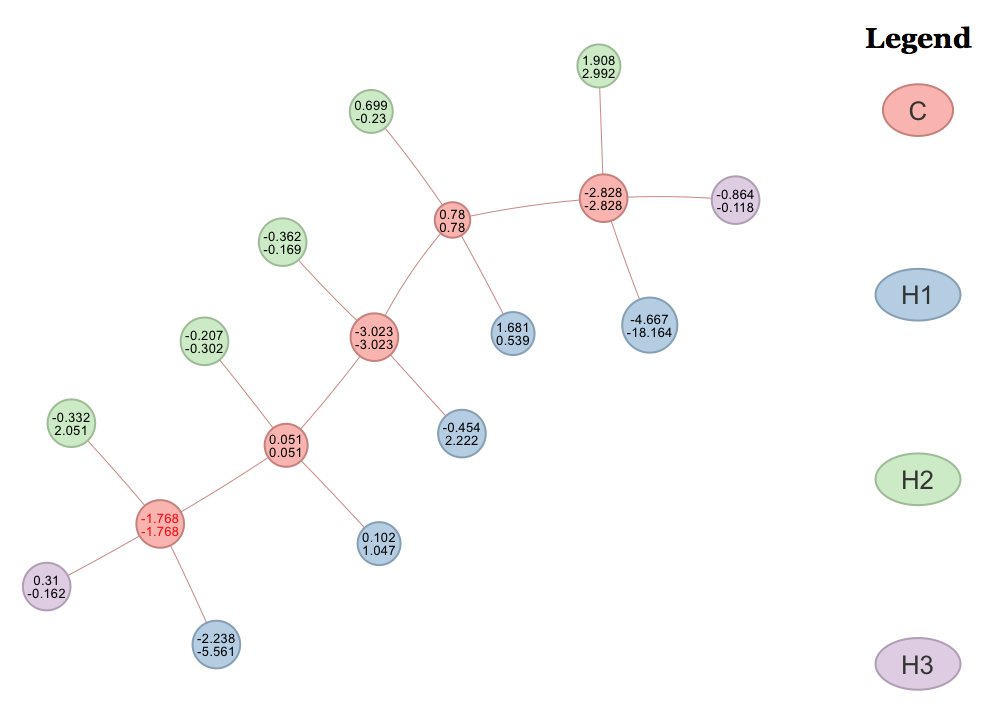}
\par\end{centering}
\caption{Visualization of the trained AHN of the nonlinear function, using the \texttt{visualize(ahn)} instruction. The network is read from left to right, starting at the carbon node with red characters. For each node, the top number corresponds to the coefficient of $x_1$, while the bottom number corresponds to the coefficient of $x_2$.\label{fig:Case2_Fig1}}
\end{figure}

\begin{figure}
\begin{centering}
\includegraphics[width=0.6\textwidth]{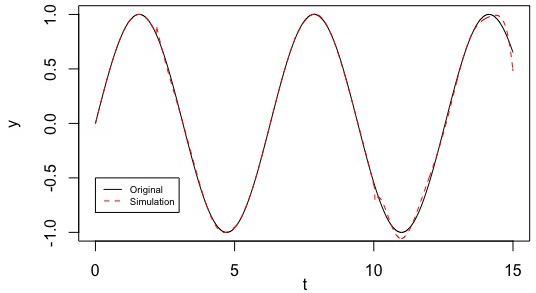}
\par\end{centering}
\caption{Function approximation using the \texttt{predict(ahn)} instruction. 
\label{fig:Case2_Fig2}}
\end{figure}

The \texttt{visualize} function of the \texttt{ahn} object is shown in Fig. \ref{fig:Case2_Fig1}; while the simulated output of it can be obtained using the \texttt{predict} function that plots Fig. \ref{fig:Case2_Fig2}. 

\subsection{Knowledge Extraction from a Dataset}

This example intends to show how to extract knowledge from a data set. We use the combined cycle power plant data set from the UCI Machine Learning Repository \cite{Yeh1998modeling}. The dataset has 9,568 observations of 5 variables. From these, only temperature and relative humidity are considered to estimate the net hourly electrical energy output, as suggested in \cite{Tufekci2014prediction}. These are summarized in Table \ref{tab:Variables Case 3}.

\begin{table}
\caption{Features used for training AHN, adapted from \cite{Tufekci2014prediction}.}
\label{tab:Variables Case 3}
\centering{ 
\footnotesize
\begin{tabular}{|l|l|l|}
\hline
\textbf{Variable}                     & \textbf{Name}                                               & \textbf{Units}                                  \\ \hline
\multirow{2}{*}{Input}  & \multicolumn{1}{l|}{Temperature}                        & \multicolumn{1}{l|}{ºC} \\ \cline{2-3}
      & \multicolumn{1}{l|}{Relative Humidity}                       & \multicolumn{1}{l|}{\%}                  \\ \hline
\multicolumn{1}{|l|}{Output} & \multicolumn{1}{l|}{ Net hourly electrical energy output} & \multicolumn{1}{l|}{MW}               \\ \hline
\end{tabular}}
\normalsize
\end{table}

The data is divided into training (70\%) and testing (30\%) sets. Furthermore, the data is standardized and the corresponding parameters are learned from the training set. We use the \texttt{fit} function to train AHN using 5 molecules, learning rate set to 0.1, and a maximum number of iterations of 200 is chosen. The code for training is shown below:

\begin{verbatim}
> power_data <- read_csv("dataset.csv")
> power_data <- as.data.frame(power_data)
> summary(power_data)

  temperature    relative_humidity  energy_power  
 Min.   : 1.81   Min.   : 992.9    Min.   :420.3  
 1st Qu.:13.51   1st Qu.:1009.1    1st Qu.:439.8  
 Median :20.34   Median :1012.9    Median :451.6  
 Mean   :19.65   Mean   :1013.3    Mean   :454.4  
 3rd Qu.:25.72   3rd Qu.:1017.3    3rd Qu.:468.4  
 Max.   :37.11   Max.   :1033.3    Max.   :495.8  

> set.seed(2018)
> sample <- sample.int(n=nrow(power_data), 
            size=floor(0.7*nrow(power_data)), replace=F)
> train_data <- power_data[sample, ]
> test_data <- power_data[-sample, ]

> library(caret)
> normalizer <- preProcess(train_data, 
                method = c("center", "scale"))
> normalized_train_data <- stats::predict(normalizer,
                train_data)

> library(ahnr)
> Sigma <- list(X=normalized_train_data[ , 1:2], 
      Y=normalized_train_data[ ,"energy_power", drop=FALSE])
> ahn <- fit(Sigma, 5, 0.1, 200)
\end{verbatim}

%
%
%
%
%
%
%
%
%
%

\begin{figure}[t]
\begin{centering}
\includegraphics[width=0.7\textwidth]{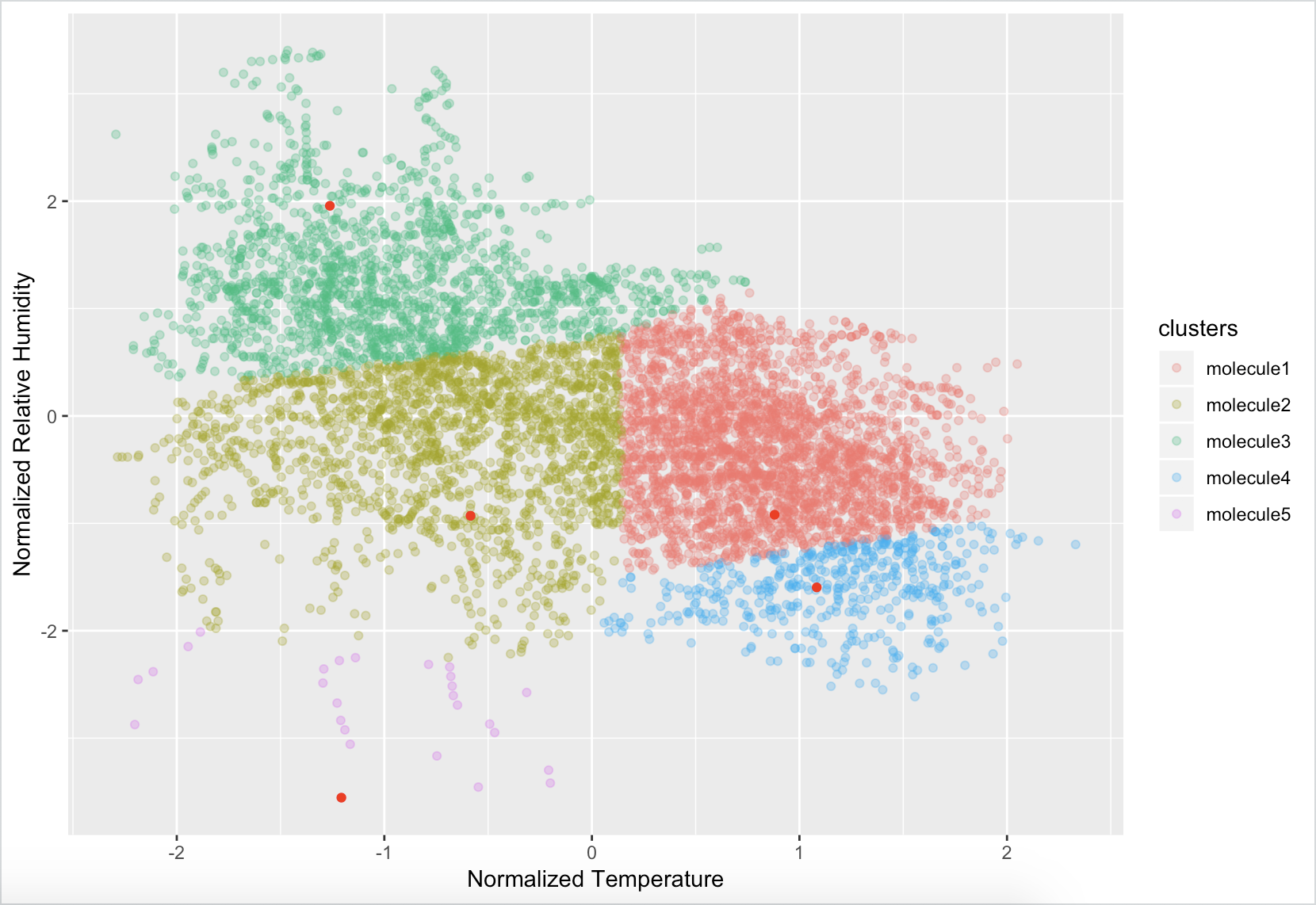}
\par\end{centering}
\caption{Training set clustered using the centers of molecules in AHN. \label{fig:Case3_Fig1}}
\end{figure}

\begin{figure}[t]
\begin{centering}
\includegraphics[width=0.7\textwidth]{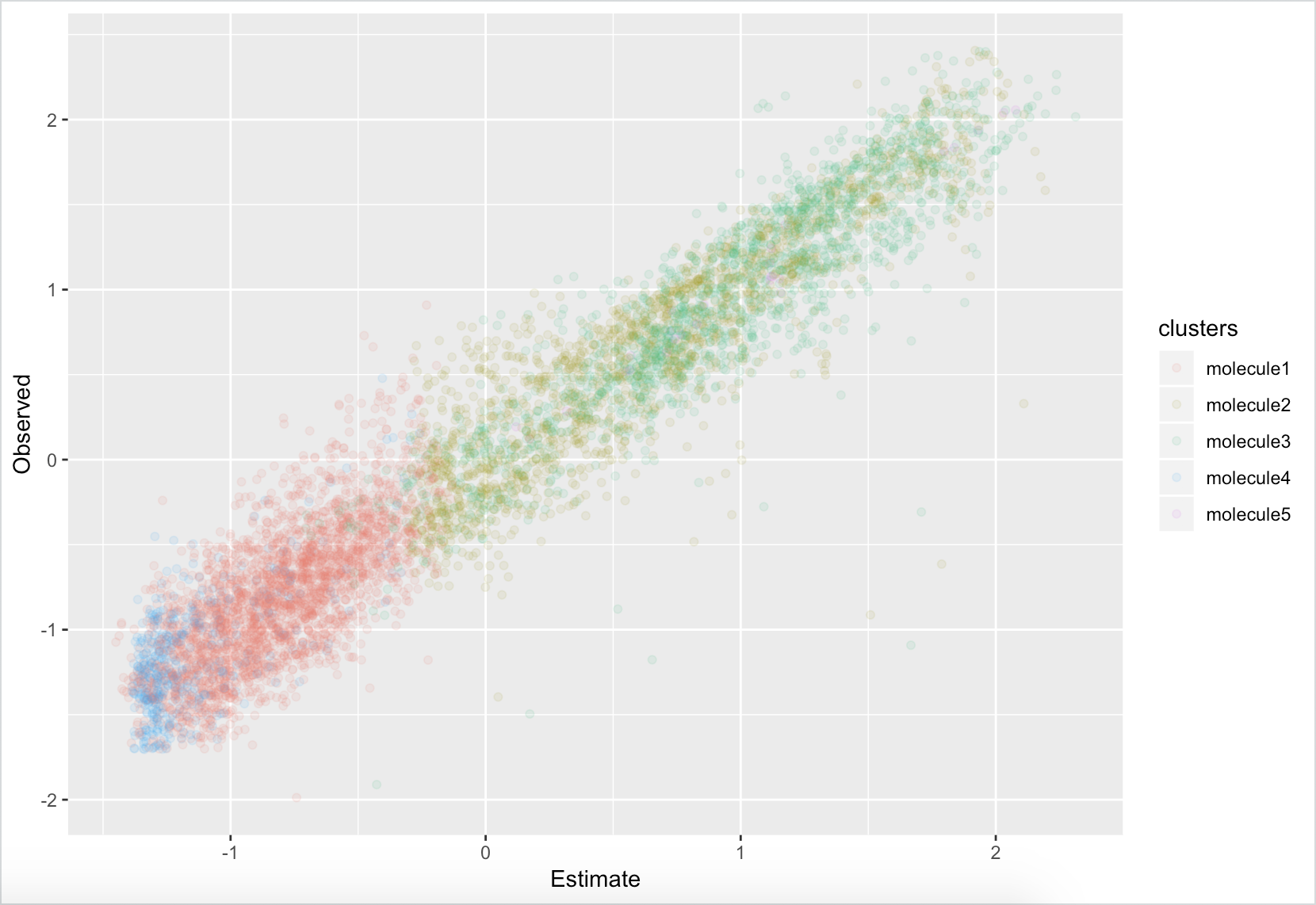}
\par\end{centering}
\caption{Comparison between the observed and predicted values for the normalized net hourly electrical energy output. \label{fig:Case3_Fig2}}
\end{figure}

In this example, we focus on the clusterization obtained by AHN. This is shown in Fig. \ref{fig:Case3_Fig1}, where the training set is segmented based on the centers got from the \texttt{summary} information. Clusters found can be applied to the simulated output of AHN and the training set, as shown in Fig. \ref{fig:Case3_Fig2}. Clusterization gives advantages on interpretability of data using AHN-model. For example, \texttt{molecule4} in Fig. \ref{fig:Case3_Fig1} represents that positive temperatures and low negative relative humidity values correspond to very low net hourly electrical energy outputs (as shown in Fig. \ref{fig:Case3_Fig2}). Other rules can be obtained in the same manner, as follows:

\begin{itemize}
    \item If temperature in negative and relative humidity is positive, then net hourly electrical energy is high.
    \item If temperature is negative and relative humidity is negative, then net hourly electrical energy is low.
    \item If temperature is negative and relative humidity is very negative, then net hourly electrical energy is very low.
\end{itemize}

\section{Prediction of Exchange Rate BRIC Currencies to USD}
\label{sec:case}

In economics, the acronymn BRIC stands for the group of countries Brazil, Russia, India and China. This is an interesting set of countries because of their rapidly growing share in international trading over the last two decades \cite{Caporale2017macro}. The BRIC group was proposed in 2001. This set of countries had an increasing global trade participation from 3\% to 19\% in the period 1990 - 2011 \cite{Caporale2017macro}. A measurement in the impact of BRIC over the global economy is to describe the exchange rate of BRIC currencies versus the U.S. dollar (USD). 

For this real-world application, modeling and prediction of the exchange rate of BRIC currencies over USD is proposed as explained below. Since data vary over time, it is important to capture the dynamics of the time series. Thus, we use a dynamic model of the series as expressed in (\ref{eq:currency}), where $y_t$ represents the exchange rate in time $t$, $f$ is a nonlinear model that captures the dynamics of the data and $\Delta y_t$ represents the difference between $y_{t-1}$ and $y_t$. Particularly, $f$ is modeled using an AHN to predict one time ahead exchange rate $y_{t+1}$.

\begin{equation}
\Delta y_{t+1} = f(y_{t-2},y_{t-1},y_t)
\label{eq:currency}
\end{equation}

\subsection{Data Preparation}
We use the public dataset about the historical data of monthly frequencies for exchange rate BRIC currencies to USD, from July 1997 to December 2015, \cite{Kaggle2018exchange}. The dataset is divided in 70\% training and 30\% testing sets. The training set consists of 156 samples representing the exchange rates, month per month, of the currencies: Brazil Real to USD (BRL/USD), Russian Ruble to USD (RUB/USD), Indian Rupee to USD (INR/USD), and Chinese Yuan to USD (YUAN/USD). Similarly, testing set consists of 65 samples in the same fashion.

For AHN, the inputs of one BRIC currency to USD are represented by $x = \{y_{t-2},y_{t-1},y_t\}$ and the output is represented by $\Delta y = y_{t+1} - y_{t}$. In addition, these values are normalized. The following code illustrates data preparation of BRL/USD:

%
%

\begin{verbatim}
> exchange_data <- as.data.frame(exchange_data)

> window = 3

> z <- matrix(0,nrow = 222, ncol=3)
> z[1,] <- exchange_data$`BRL/USD`[1:3]
> z[2,] <- exchange_data$`BRL/USD`[2:4]
> z[3,] <- exchange_data$`BRL/USD`[3:5]

> y <- matrix(0, nrow=222, ncol=1)

> for (i in 1:(222-window)){
>  z[i+window,] <- exchange_data$`BRL/USD`[(i+1):(i+window)]
>  y[i+window,] <- z[i+window,3] - z[i+window-1,3]
> }

> x <- z[1:221,]
> y <- y[2:222,]

> train_x_data <- as.data.frame(x[1:156,])
> train_y_data <- as.data.frame(y[1:156])

> test_x_data <- as.data.frame(x[157:221,])
> test_y_data <- as.data.frame(y[157:221])

> library(caret)
> set.seed(201708)

> normalizer_x <- preProcess(train_x_data, 
                method = c("center", "scale"))
> train_xn     <- stats::predict(normalizer_x, train_x_data)

> normalizer_y <- preProcess(train_y_data, 
                method = c("center", "scale"))
> train_yn     <- stats::predict(normalizer_y, train_y_data)
\end{verbatim}

\subsection{Training Step}
Using the training set prepared previously, the AHN is trained. For this purpose, we use 10 molecules, 0.1 in learning rate and 2000 of maximum iterations. From (\ref{eq:currency}), the predicted value $y_{t+1}$ can be obtained as expressed in (\ref{eq:solution}), where $\hat{f}$ is estimated using the AHN model, i.e. $\hat{f}\sim AHN$. 

\begin{equation}
\hat{y}_{t+1} = y_t + \hat{f}(y_{t-2},y_{t-1},y_t)
\label{eq:solution}
\end{equation}

The AHN-model obtained an overall error $0.9014$. Figure \ref{fig:ahn-real} shows the visualization of the trained AHN. Once the predicted values are computed using AHN, the mean-squared error (MSE) calculated was $0.0147$. The training step is shown following:


\begin{verbatim}
> library(ahnr)
> Sigma <- list(X=train_xn, Y=train_yn[,1, drop=FALSE])
> ahn <- fit(Sigma, 10, 0.1, 2000)
\end{verbatim}

\begin{figure}
\begin{centering}
\includegraphics[width=0.7\textwidth]{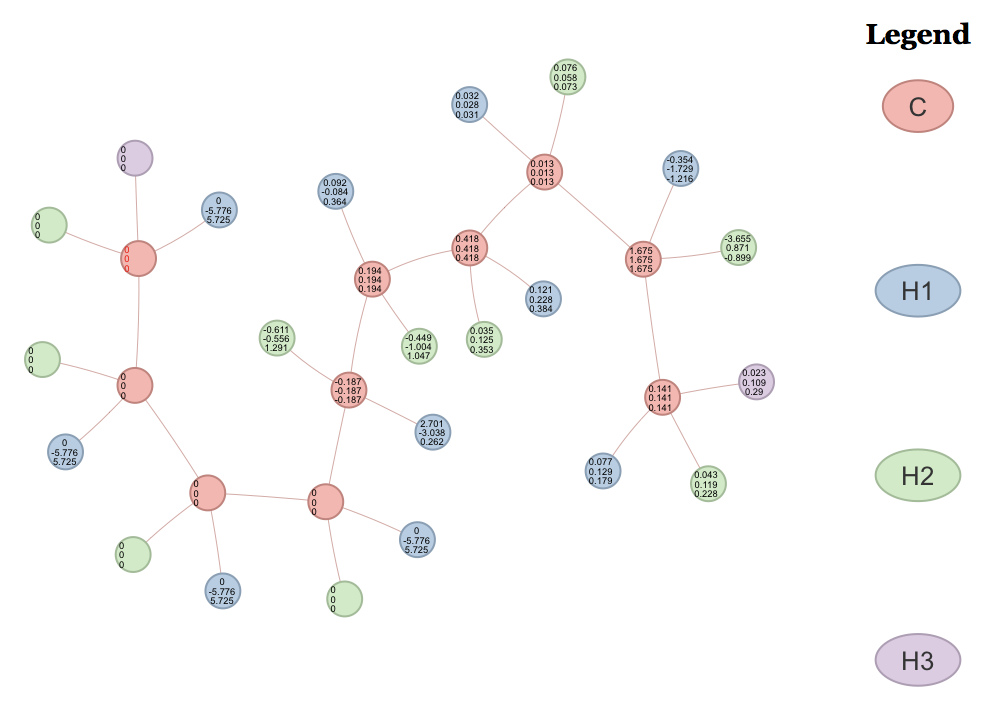}
\par\end{centering}
\caption{AHN-model for BRL/USD. For each node, the top number corresponds to the coefficient of $y_t$, next is the coefficient of $y_{t-1}$ while the bottom number corresponds to the coefficient of $y_{t-2}$.  \label{fig:ahn-real}}
\end{figure}

\subsection{Testing Step}
\begin{figure}
\begin{centering}
\includegraphics[width=0.7\textwidth]{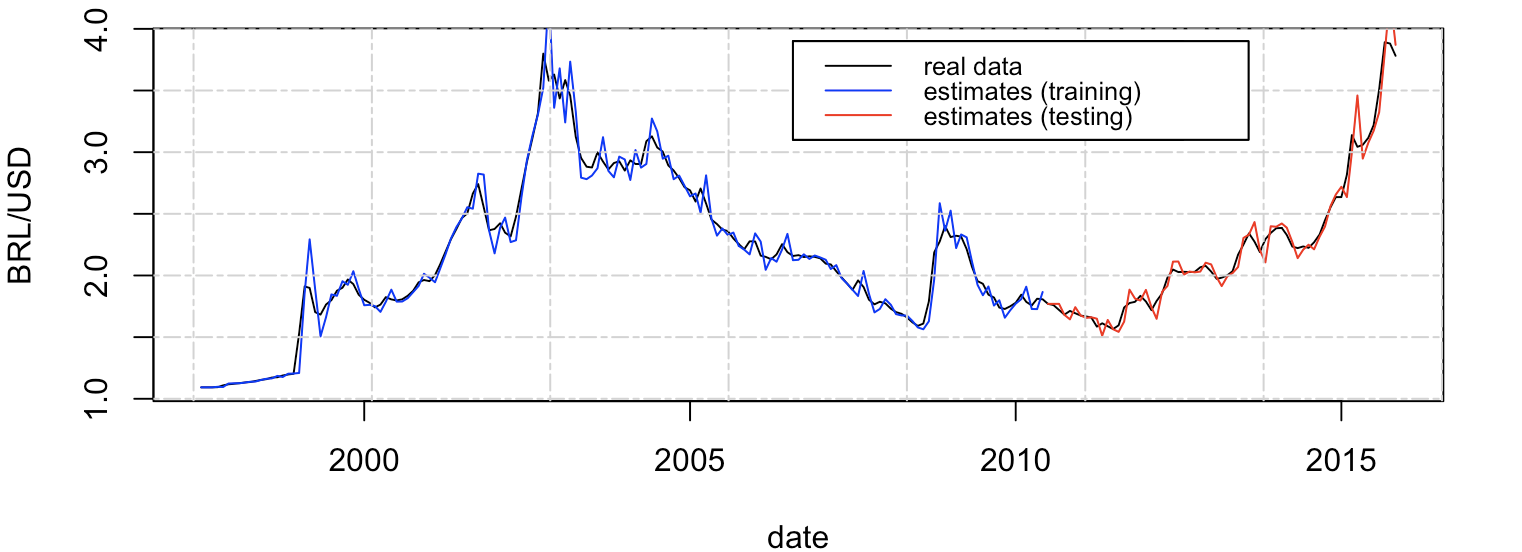}
\par\end{centering}
\caption{Prediction response of BRL/USD using the AHN-model. Original data in black line, estimates of training values in blue line and estimates of testing values in red line. \label{fig:graph-real}}
\end{figure}

Using the testing set and the trained AHN, the predicted exchange rate values $y_{t+1}$ were computed using (\ref{eq:solution}). The MSE calculated was $0.0102$. Figure \ref{fig:graph-real} shows the estimates of BRL/USD in comparison with the real time series (black line). The estimates calculated in the training set range are shown in blue and the ones in the testing set range are shown in red. The following code summarizes the prediction of BRL/USD:

\begin{verbatim}
> test_xn      <- predict(normalizer_x, test_x_data)

> y_prediction <- matrix(0, nrow=65, ncol=1)
> for (i in 2:65){
>     tmp <- test_xn[i-1,]
>     delta <- predict(ahn,tmp) * normalizer_y$std[1] 
             + normalizer_y$mean[1]
>     y_prediction[i,1] <- delta + test_x_data[i-1,3]
> }
\end{verbatim}

\subsection{Predictions of BRIC}
\begin{figure}
\begin{centering}
\includegraphics[width=0.7\textwidth]{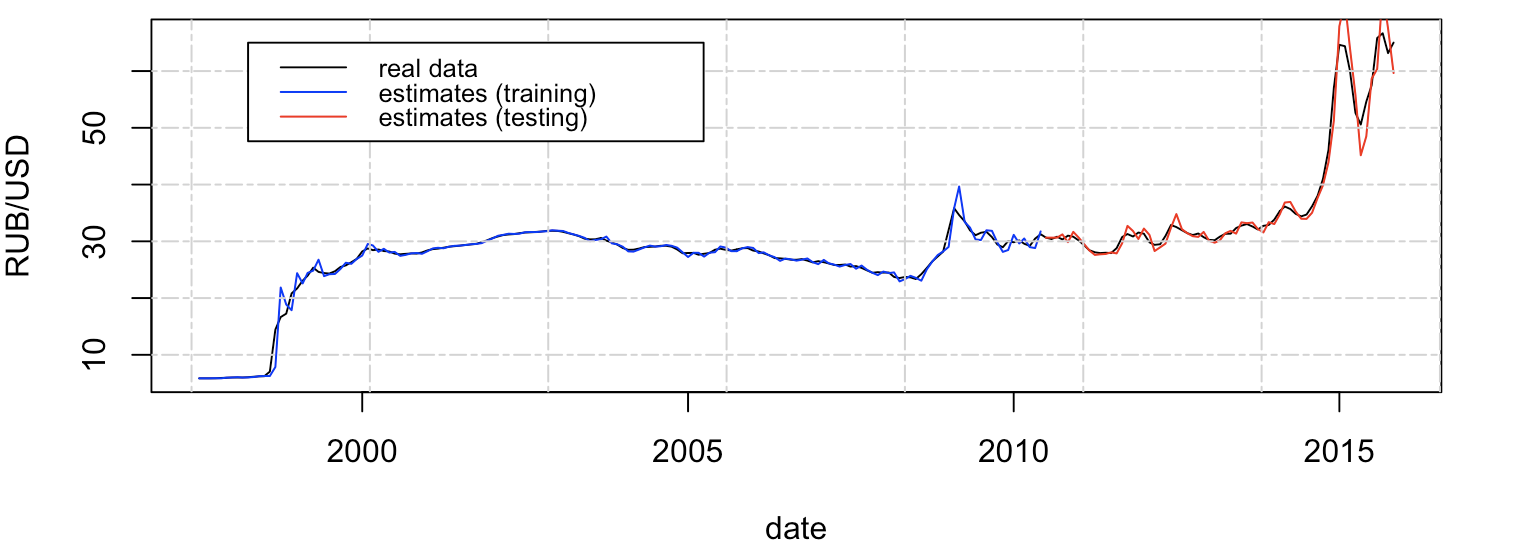}\\
\includegraphics[width=0.7\textwidth]{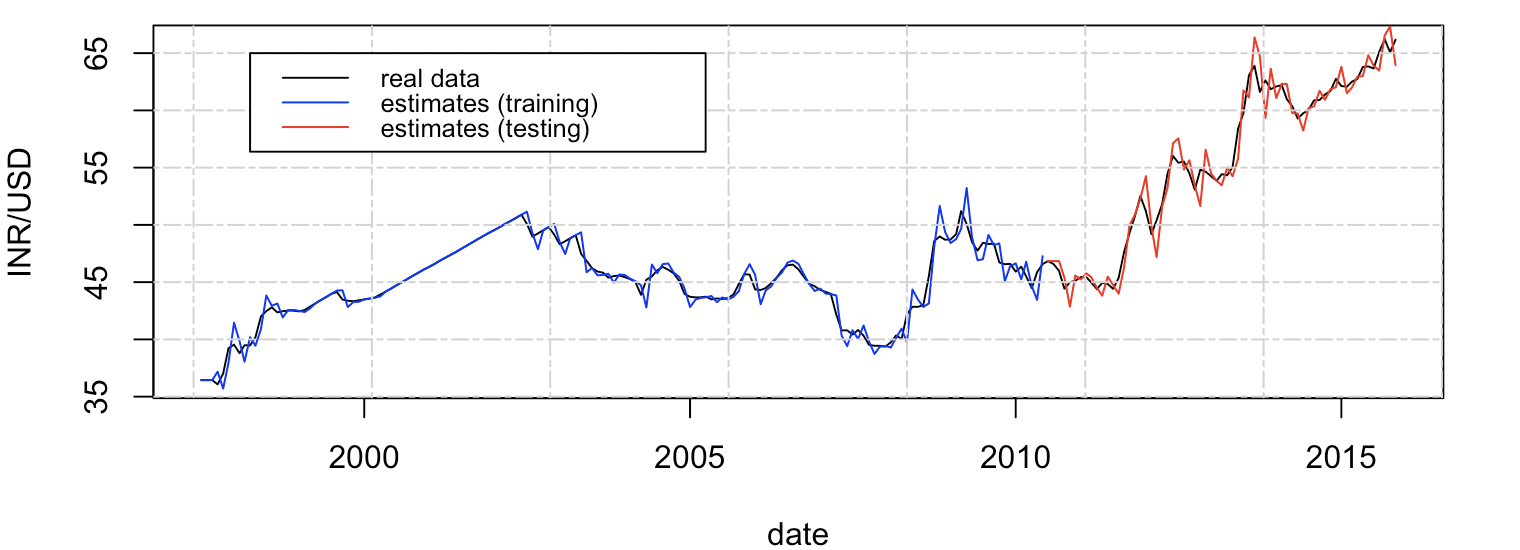}\\
\includegraphics[width=0.7\textwidth]{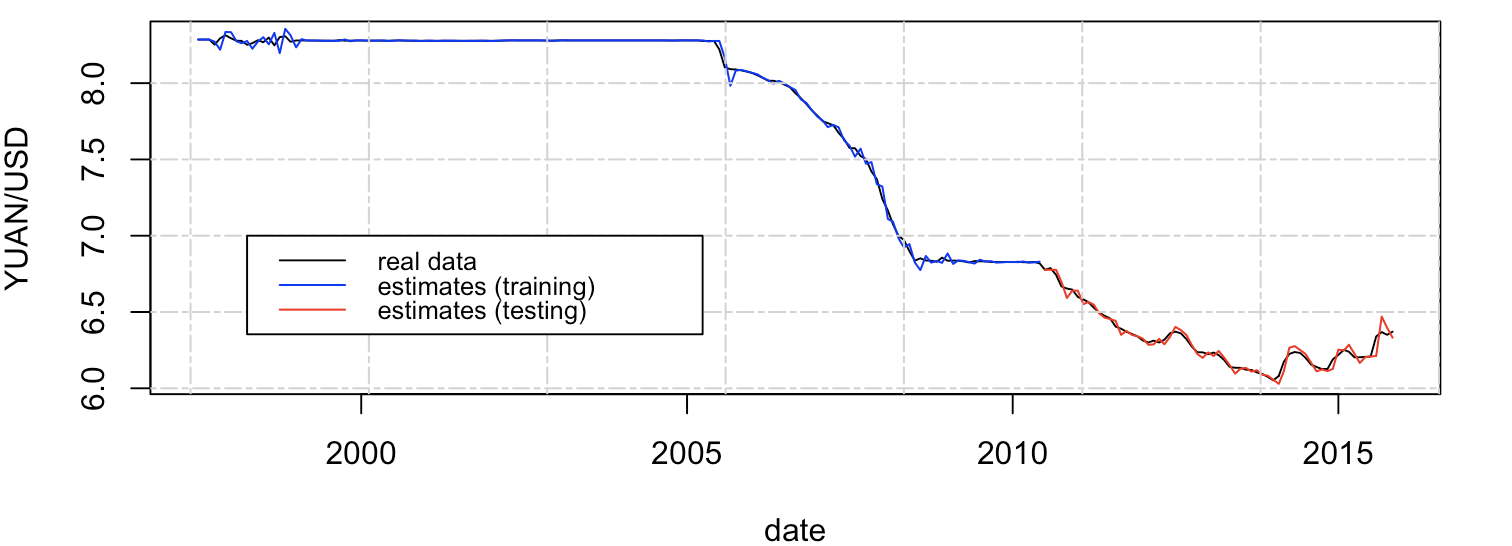}
\par\end{centering}
\caption{Prediction response of (top-to-bottom): RUB/USD, INR/USD and YUAN/USD using AHN. Original data in black line, estimates of training values in blue line and estiamtes of testing values in red line. \label{fig:russian}}
\end{figure}

Moreover, the same workflow was implemented for the other three exchange rate currencies over USD. Figure \ref{fig:russian} shows the corresponding prediction responses of RUB/USD, INR/USD and YUAN/USD, from top to bottom, using a trained AHN for each exchange rate. The same parameters for training were employed: 10 molecules, 0.1 in learning rate and 2000 as maximum iterations; except for YUAN/USD which it was trained with 5 molecules.

%

\section{Limitations}
\label{sec:limitations}

Although AHN is a promising data-modeling technique, there are a couple of issues that require further research. The first one is how to select the number of molecules. This hyperparameter is important because it defines the complexity of the network. A small number of molecules will not be able to achieve a good approximation, while a large number will result in an unnecessary use of computational resources. In \cite{Ponce2014artificial}, there is a proposed heuristic to automatically tune this hyperparameter, but more analysis should be done on this before being implemented. Hence, we suggest to use cross-validation or Bayesian optimization to find suitable hyperparameters.

Also, the current implementation of the algorithm is susceptible to get stuck in local minima during training, and as the number of input variables increases, the technique suffers from the curse of dimensionality. Furthermore, the package can only train a single network or component at a time, although it has been shown that several networks can be trained at once to obtain a better input-output model \cite{Ponce2014artificial}.  These issues are planned to be addressed in future releases of the package.

\section{Conclusions}
\label{sec:conclusion}

This article describes the \texttt{ahnr} package that allows the creation, training, testing and visualization of artificial hydrocarbon networks. Two examples and one case study were presented to show the usage of the package. 
 
Particularly, the examples showed how to approximate nonlinear functions, and how to extract knowledge using the centers of molecules and a correlation to the estimates. Moreover, these examples show the effectiveness and the easiness of AHN implementation with the package. In addition, we demonstrated the predictive power of AHN using a public and real dataset of exchange rates from the BRIC currencies over USD. We showed the step-by-step implementation of AHN using the \texttt{ahnr} package and how to capture the dynamics of a time series. As shown in the results, all BRIC currencies to USD exchange rates were predicted successfully. 

For future work, we will expand the \texttt{ahnr} package to tackle some of the limitations reported above, and we will welcome feedback from the users.

\bibliographystyle{unsrt}
\bibliography{article}

\begin{thebibliography}{10}

\bibitem{Ponce2011artificial}
Hiram Ponce and Pedro Ponce.
\newblock Artificial organic networks.
\newblock In {\em Electronics, Robotics and Automotive Mechanics Conference
  (CERMA)}, pages 29--34. IEEE, 2011.

\bibitem{Ponce2014artificial}
Hiram Ponce, Pedro Ponce, and Arturo Molina.
\newblock {\em Artificial Organic Networks: Artificial Intelligence Based on
  Carbon Networks}, volume 521 of {\em Studies in Computational Intelligence}.
\newblock Springer, 2014.

\bibitem{Ponce2016flexible}
Hiram Ponce, Luis Miralles-Pechu\'an, and Lourdes Mart\'inez-Villasenor.
\newblock A flexible approach for human activity recognition using artificial
  hydrocarbon networks.
\newblock {\em Sensors}, 16(11):1715, 2016.

\bibitem{Ponce2015artificial}
Hiram Ponce, Luis Miralles-Pechu\'an, and Lourdes Mart\'inez-Villasenor.
\newblock Artificial hydrocarbon networks for online sales prediction.
\newblock In {\em Mexican International Conference on Artificial Intelligence},
  volume 9414, pages 498 -- 508. Springer, 2015.

\bibitem{Ponce2014adaptive}
Hiram Ponce, Pedro Ponce, and Arturo Molina.
\newblock Adaptive noise filtering based on artificial hydrocarbon networks: An
  application to audio signals.
\newblock {\em Expert Systems With Applications}, 41(14):6512--6523, 2014.

\bibitem{Ponce2013artificial}
Hiram Ponce, Pedro Ponce, and Arturo Molina.
\newblock Artificial hydrocarbon networks fuzzy inference system.
\newblock {\em Mathematical Problems in Engineering}, 2013(531031):1--13, 2013.

\bibitem{Ponce2015novel}
Hiram Ponce, Pedro Ponce, H\'ector Bastida, and Arturo Molina.
\newblock A novel robust liquid level controller for coupled-tanks system using
  artificial hydrocarbon networks.
\newblock {\em Expert Systems With Applications}, 42(22):8858 -- 8867, 2015.

\bibitem{Ponce2017doubly}
Pedro Ponce, Hiram Ponce, and Arturo Molina.
\newblock Doubly fed induction generator ({DFIG}) wind turbine controlled by
  artificial organic networks.
\newblock {\em Soft Computing}, pages 1--13, 2017.

\bibitem{Ponce2015development}
Hiram Ponce, Pedro Ponce, and Arturo Molina.
\newblock The development of an artificial organic networks toolkit for
  labview.
\newblock {\em Journal of Computational Chemistry}, 36(7):478--492, 2015.

\bibitem{Dogaru2015using}
Ioana Dogaru and Radu Dogaru.
\newblock Using {Python and Julia} for efficient implementation of natural
  computing and complexity related algorithms.
\newblock In {\em 20th International Conference on Control Systems and Computer
  Science}, pages 599 -- 604, 2015.

\bibitem{Lantz2015machine}
Brett Lantz.
\newblock {\em Machine Learning with {R}}.
\newblock Packt Publisher, 2015.

\bibitem{Ponce2016novel}
Hiram Ponce, Lourdes Mart\'inez-Villasenor, and Luis Miralles-Pechu\'an.
\newblock A novel wearable sensor-based human activity recognition approach
  using artificial hydrocarbon networks.
\newblock {\em Sensors}, 16(7):1033, 2016.

\bibitem{Ponce2017interpretability}
Hiram Ponce and Lourdes Mart\'inez-Villasenor.
\newblock Interpretability of artificial hydrocarbon networks for breast cancer
  classification.
\newblock In {\em 30th International Joint Conference on Neural Networks},
  pages 3535--3542. IEEE, 2017.

\bibitem{Yeh1998modeling}
I-Cheng Yen.
\newblock Modeling of strength of high performance concrete using artificial
  neural networks.
\newblock {\em Cement and Concrete Research}, 28(12):1797--1808, 1998.

\bibitem{Tufekci2014prediction}
Pinar Tufekci.
\newblock Prediction of full load electrical power output of a base load
  operated combined cycle power plant using machine learning methods.
\newblock {\em International Journal of Electrical Power \& Energy Systems},
  60(9):126--140, 2014.

\bibitem{Caporale2017macro}
Guglielmo~M. Caporale, Fabio Spagnolo, and Nicola Spagnolo.
\newblock Macro news and exchange rates in the {BRICS}.
\newblock {\em Finance Research Letters}, 21(5):140--143, 2017.

\bibitem{Kaggle2018exchange}
Kaggle-Dataset.
\newblock Exchange rate {BRIC currencies/US dollar}, 2018.

\end{thebibliography}

\end{document}